%% file: Camera_Ready.tex
\documentclass[10pt,twocolumn,letterpaper]{article}

\usepackage{cvpr}              

\usepackage{graphicx}
\usepackage{amsmath}
\usepackage{amssymb}
\usepackage{booktabs}

\usepackage{multirow}
\usepackage[ruled]{algorithm2e}

\usepackage[accsupp]{axessibility}

\usepackage[pagebackref,breaklinks,colorlinks]{hyperref}

\usepackage[capitalize]{cleveref}
\crefname{section}{Sec.}{Secs.}
\Crefname{section}{Section}{Sections}
\Crefname{table}{Table}{Tables}
\crefname{table}{Tab.}{Tabs.}

\def\cvprPaperID{18} 
\def\confName{CVPR}
\def\confYear{2023}

\begin{document}

\title{Are Data-driven Explanations Robust against Out-of-distribution Data?}

\author{Tang Li \qquad Fengchun Qiao \qquad Mengmeng Ma \qquad Xi Peng\\
University of Delaware\\
{\tt\small \{tangli, fengchun, mengma, xipeng\}@udel.edu}
}
\maketitle

\begin{abstract}
As black-box models increasingly power high-stakes applications, a variety of data-driven explanation methods have been introduced.
Meanwhile, machine learning models are constantly challenged by distributional shifts.
A question naturally arises: Are data-driven explanations robust against out-of-distribution data?
Our empirical results show that even though predict correctly, the model might still yield unreliable explanations under distributional shifts.
How to develop robust explanations against out-of-distribution data?
To address this problem, we propose an end-to-end model-agnostic learning framework Distributionally Robust Explanations (DRE).
The key idea is, inspired by self-supervised learning, to fully utilizes the inter-distribution information to provide supervisory signals for the learning of explanations without human annotation.
Can robust explanations benefit the model’s generalization capability?
We conduct extensive experiments on a wide range of tasks and data types, including classification and regression on image and scientific tabular data.
Our results demonstrate that the proposed method significantly improves the model's performance in terms of explanation and prediction robustness against distributional shifts.
\end{abstract}

\section{Introduction}
\label{sec:intro}
\input{Section_Camera_ready/1_Introduction.tex}


\section{Related work}
\label{sec:relate}
\input{Section_Camera_ready/2_Related_work.tex}

\section{Methods}
\label{sec:method}
\input{Section_Camera_ready/3_Method.tex}

\section{Experiments}
\label{sec:exp}
\input{Section_Camera_ready/4_Experiments.tex}

\section{Conclusion}
\label{sec:conclusion}
\input{Section_Camera_ready/5_Conclusion.tex}

\section*{Acknowledgement}
We would like to thank the anonymous reviewers for their insightful feedback.
This work is partially supported by the General University Research (GUR) and the University of Delaware Research Foundation (UDRF).

{\small
\bibliographystyle{ieee_fullname}
\bibliography{egbib}
}

\end{document}

%% file: Section_Camera_ready/1_Introduction.tex
\begin{figure}[t]
\centering
   \includegraphics[width=1.0\linewidth]{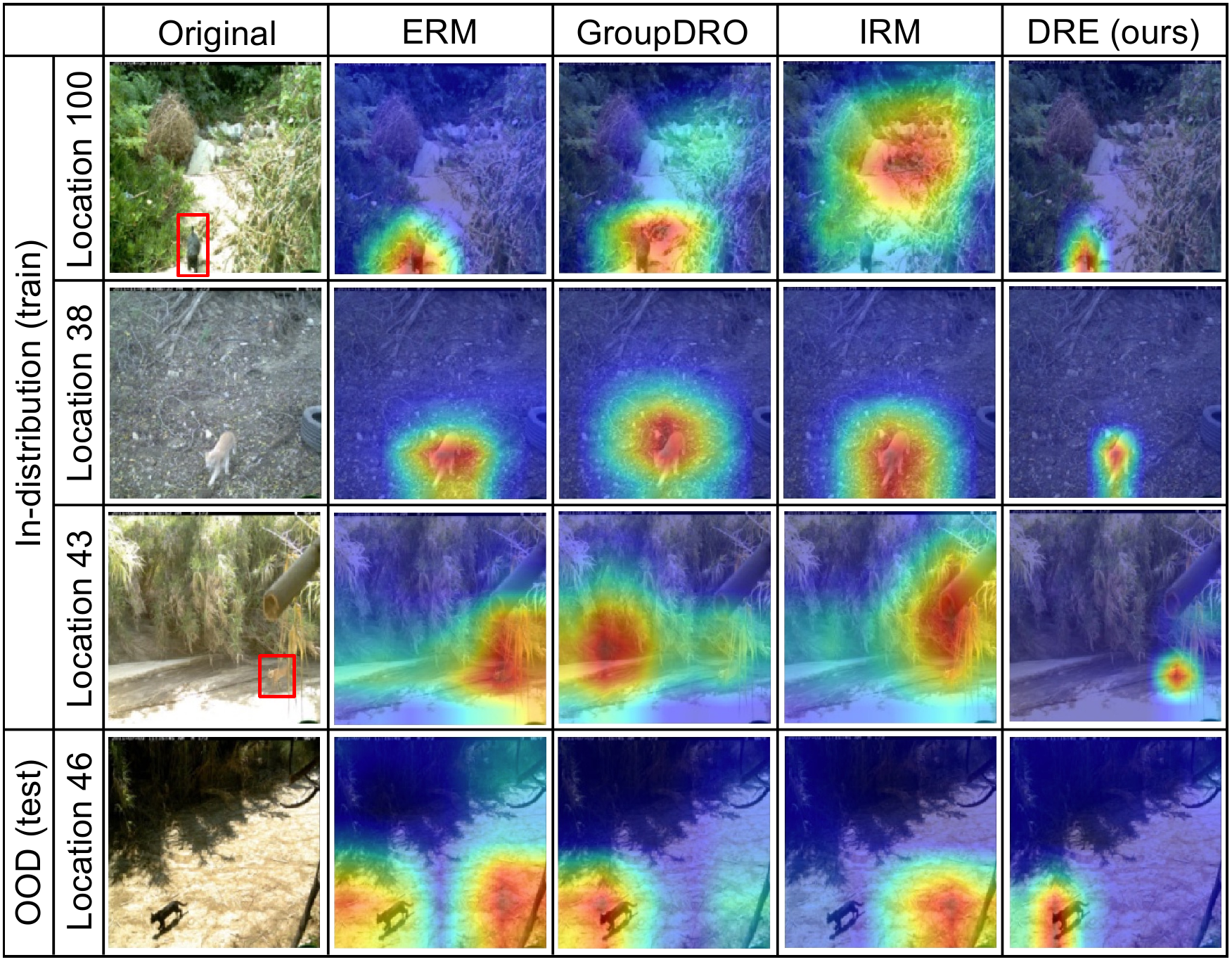}

\caption{
The explanations for {\it in-} and {\it out-of-distribution} data of {\it Terra Incognita}~\cite{beery2018recognition} dataset. 
Note that GroupDRO~\cite{sagawa2019distributionally} and IRM~\cite{arjovsky2019invariant} are explicitly designed methods that can predict accurately across distributions.
Although with correct predictions, the explanations of models trained by such methods would also highlight distribution-specific associations ({\it e.g.}, tree branches) except the object. 
This leads to unreliable explanations on OOD data.
On the contrary, our model consistently focuses on the most discriminative features shared across distributions.
}
\label{fig:intro}
\end{figure}

\footnotetext[1]{The source code and pre-trained models are available at: \url{https://github.com/tangli-udel/DRE}.}

There has been an increasing trend to apply {\it black-box} machine learning (ML) models for high-stakes applications.
The lack of explainability of models can have severe consequences in healthcare~\cite{rudin2019stop}, criminal justice~\cite{wexler2017computer}, and other domains.
Meanwhile, ML models are inevitably exposed to unseen distributions that lie outside their training space~\cite{torralba2011unbiased, lake2017building};
a highly accurate model on average can fail catastrophically on {\it out-of-distribution} (OOD) data due to naturally-occurring variations, sub-populations, spurious correlations, and adversarial attacks.
For example, a cancer detector would erroneously predict samples from hospitals having different data acquisition protocols or equipment manufacturers.
Therefore, reliable explanations across distributions are crucial for the safe deployment of ML models.
However, existing works focus on the reliability of data-driven explanation methods~\cite{adebayo2018sanity, zhou2022feature} while ignoring the robustness of explanations against distributional shifts.

A question naturally arises: {\it Are data-driven explanations robust against out-of-distribution data?}
We empirically investigate this problem across different methods. 
Results of the Grad-CAM~\cite{selvaraju2017grad} explanations are shown in Fig.~\ref{fig:intro}.
We find that {\it the distributional shifts would further obscure the decision-making process due to the black-box nature of ML models}.
As shown, the explanations focus not only on the object but also spurious factors ({\it e.g.}, background pixels).
Such distribution-specific associations would yield {\it inconsistent} explanations across distributions.
Eventually, it leads to unreliable explanations ({\it e.g.}, tree branches) on OOD data.
This contradicts with human prior that the most discriminative features ought to be invariant.

{\it How to develop robust explanations against out-of-distribution data?}
Existing works on OOD generalization are limited to data augmentation~\cite{volpi2018generalizing, qiao2020learning, shankar2018generalizing}, distribution alignment~\cite{ganin2016domain, li2018domain, bahng2020learning}, Meta learning~\cite{li2018learning, dou2019domain, qiao2021uncertainty}, or invariant learning~\cite{arjovsky2019invariant, krueger2021out}.
However, without constraints on explanations, the model would still recklessly absorb all associations found in the training data, including {\it spurious correlations}.
To constrain the learning of explanations, existing methods rely on explanation annotations~\cite{rieger2020interpretations, stammer2021right} or one-to-one mapping between image transforms~\cite{guo2019visual, pillai2022consistent, chen2019robust}.
However, there is no such mapping in general {\it naturally-occurring} distributional shifts.
Furthermore, obtaining ground truth explanation annotations is prohibitively expensive~\cite{wang2020self}, or even impossible due to subjectivity in real-world tasks~\cite{roscher2020explainable}.
To address the aforementioned limitations, we propose an end-to-end model-agnostic training framework {\it Distributionally Robust Explanations (DRE)}.
The key idea is, inspired by self-supervised learning, to fully utilize the inter-distribution information to provide supervisory signals for explanation learning.

{\it Can robust explanations benefit the model's generalization capability?}
We evaluate the proposed methods on a wide range of tasks in Sec.~\ref{sec:exp}, including the classification and regression tasks on image and scientific tabular data.
Our empirical results demonstrate the robustness of our explanations.
The explanations of the model trained via the proposed method outperform existing methods in terms of explanation consistency, fidelity, and scientific plausibility.
The extensive comparisons and ablation studies prove that our robust explanations significantly improve the model's prediction accuracy on OOD data.
As shown, the robust explanations would alleviate the model's excessive reliance on {\it spurious correlations}, which are unrelated to the causal correlations of interest~\cite{arjovsky2019invariant}.
Furthermore, the enhanced explainability can be generalized to a variety of data-driven explanation methods.

In summary, our main contributions:
\begin{itemize}
  \item We comprehensively study the robustness of data-driven explanations against naturally-occurring distributional shifts.

  \item We propose an end-to-end model-agnostic learning framework Distributionally Robust Explanations (DRE). 
  It fully utilizes inter-distribution information to provide supervisory signals for explanation learning without human annotations.

  \item Empirical results in a wide range of tasks including classification and regression on image and scientific tabular data demonstrate superior explanation and prediction robustness of our model against OOD data.
\end{itemize}

%% file: Section_Camera_ready/2_Related_work.tex
{\bf Explainable machine learning.}
A suite of techniques has been proposed to reveal the decision-making process of modern {\it black-box} ML models.
One direction is intrinsic to the model design and training, rendering an explanation along with its output, {\it e.g.}, attention mechanisms~\cite{vaswani2017attention} and joint training~\cite{hind2019ted, chen2019looks}.
A more popular way is to give insight into the learned associations of a model that are not readily interpretable by design, known as post-hoc methods.
Such methods usually leverage backpropagation or local approximation to offer saliency maps as explanations, {\it e.g.}, Input Gradient~\cite{simonyan2013deep}, Grad-CAM~\cite{selvaraju2017grad}, LIME~\cite{ribeiro2016should}, and SHAP~\cite{lundberg2017unified}.
Recent works have shed light on the downsides of post-hoc methods.
The gradient-based explanations ({\it e.g.}, Input Gradient) are consistent with sample-based explanations ({\it e.g.}, LIME) with comparable fidelity but have much lower computational cost~\cite{ross2017right}.
Moreover,  only the Input Gradient and Grad-CAM methods passed the sanity checks in~\cite{adebayo2018sanity}.
In our work, we incorporate gradient-based methods into optimization to calculate explanations efficiently.

{\bf Out-of-distribution generalization.}
To generalize machine learning models from training distributions to unseen distributions, existing methods on OOD generalization can be categorized into four branches:
(1) Data augmentation. \cite{volpi2018generalizing, peng2018jointly, shankar2018generalizing, ma2022multimodal} enhance the generalization performance by increasing the diversity of data through augmentation.
(2) Distribution alignment. \cite{li2018domain, bahng2020learning, qiao2023topology} align the features across source distributions in latent space to minimize the distribution gaps.
(3) Meta learning. \cite{li2018learning, dou2019domain, peng2022out, ma2021smil} using meta-learning to facilitate fast-transferable model initialization.
(4) Invariant learning. \cite{arjovsky2019invariant, krueger2021out, peng2017reconstruction} learn invariant representations that are general and transferable to different distributions.
However, recent works~\cite{gulrajani2020search, koh2021wilds} show that the classic empirical risk minimization (ERM)~\cite{vapnik1999nature} method has comparable or even outperforms the aforementioned approaches.
We argue that this is because the existing approaches barely have constraints on explanations, the model would still recklessly absorb any correlations identified in the training data.

{\bf Explanation-guided learning.}
Several recent works have attempted to incorporate explanations into model training to improve predictive performance.
\cite{rieger2020interpretations, stammer2021right} match explanations with human annotations based on domain knowledge, to alleviate the model's reliance on background pixels.
\cite{guo2019visual, wang2020self, pillai2022consistent} align explanations between spatial transformations to improve image classification and weakly-supervised segmentation.
\cite{chen2019robust, han2021explanation, cugu2022attention} synchronize the explanations of the perturbed and original samples to enhance the robustness of models.
However, acquiring ground truth explanations is prohibitively labor-intensive~\cite{wang2020self} or even impossible due to subjectivity in real-world tasks~\cite{roscher2020explainable}.
Furthermore, image transformations are insufficient to address different data types and the general {\it naturally-occurring} distributional shifts, there is no one-to-one correlation between samples from different distributions to provide supervision.


%% file: Section_Camera_ready/3_Method.tex
\subsection{Robustness against Out-of-distribution Data}
Question: {\it Are data-driven explanations robust against out-of-distribution data?}

We empirically investigate this problem on an out-of-distribution generalization benchmark image dataset {\it Terra Incognita}~\cite{beery2018recognition} and a scientific tabular dataset {\it Urban Land}~\cite{gao2020mapping}.
For image data, we leverage the Grad-CAM~\cite{selvaraju2017grad} method to generate explanations, and leverage explanation fidelity~\cite{Petsiuk2018rise} as the metric to measure explanation quality.
Specifically, we evaluate ERM~\cite{vapnik1999nature} and two representative out-of-distribution generalization methods, GroupDRO~\cite{sagawa2019distributionally} and IRM~\cite{arjovsky2019invariant}.
For scientific tabular data, we leverage the Input Gradient~\cite{simonyan2013deep} method to generate explanations, and leverage scientific consistency (Sec.~\ref{sec:metric}) as the metric to measure explanation quality.
Qualitatively, the more explanation plausibility on OOD data, the better the explanation robustness.
Quantitatively, the higher the explanation fidelity or scientific consistency on OOD data, the better the explanation robustness.

For qualitative evaluation, we observe that even with correct predictions, the explanations on out-of-distribution data might still focus on spurious correlations. 
Fig.~\ref{fig:intro} shows the Grad-CAM explanations of models trained via ERM, GroupDRO, and IRM.
As shown, the explanations on in-distribution data would not only highlight the object, but also distribution-specific associations ({\it e.g.}, background pixels).
This eventually leads to unreliable explanations on OOD data ({\it e.g.}, tree branches).
We find that the out-of-distribution generalization methods perform even worse than the classic ERM in terms of explanation robustness.
This finding verify the results in \cite{gulrajani2020search, koh2021wilds} that ERM outperforms the out-of-distribution generalization methods from an explanation perspective.
For quantitative evaluation, we empirically verify that the explanation fidelity and scientific consistency severely dropped on OOD data.
Tab.~\ref{tab:question_1} reports the evaluation results, as shown, on the average of each distribution as the testing set, the explanation fidelity dropped by 24.9\%, 20.8\%, and 15.0\% for ERM, GroupDRO, and IRM on {\it Terra Incognita};
the scientific consistency dropped by 32.4\% for ERM on {\it Urban Land}.

To conclude, {\it data-driven explanations are not robust again out-of-distribution data.}
The distributional shifts further obscure the decision-making process due to the {\it black-box} nature of modern ML models.

\begin{table}[t]
\centering
\resizebox{1.0\columnwidth}{!}{%
\begin{tabular}{llccc}
\toprule[1pt]
\multirow{2}{*}{Dataset}         & \multirow{2}{*}{Method} & \multicolumn{2}{c}{Evaluation*} & \multirow{2}{*}{$\Delta$ $\downarrow$} \\ \cline{3-4}
                                 &                         & ID             & OOD            &                                        \\ \toprule[1pt]
\multirow{3}{*}{\begin{tabular}[c]{@{}l@{}}Terra\\ Incognita~\cite{beery2018recognition}\end{tabular}} & ERM~\cite{vapnik1999nature}                     & 0.778          & 0.584          & {\bf 0.194}                                  \\
                                 & GroupDRO~\cite{sagawa2019distributionally}                & 0.742          & 0.597          & 0.145                                  \\
                                 & IRM~\cite{arjovsky2019invariant}                     & 0.612          & 0.520          & 0.092                                  \\ \toprule[0.5pt]
Urban Land~\cite{gao2020mapping}                       & ERM~\cite{vapnik1999nature}                     & 0.535          & -0.113         & {\bf 0.648}                                  \\ \bottomrule[1pt]
\end{tabular}
}
\caption{
Evaluation of the explanation quality of {\it in-distribution} (ID) and {\it out-of-distribution} data.
The results are on the average of each distribution as the testing set.
Note that the explanation quality of both image and scientific data severely dropped on OOD data.
*The reported evaluation scores are explanation fidelity~\cite{Petsiuk2018rise} ({\it Terra Incogenita}) and scientific consistency ({\it Urban Land}).
}
\label{tab:question_1}
\end{table}

\begin{figure*}[t]
\centering
\includegraphics[width=0.9\linewidth]{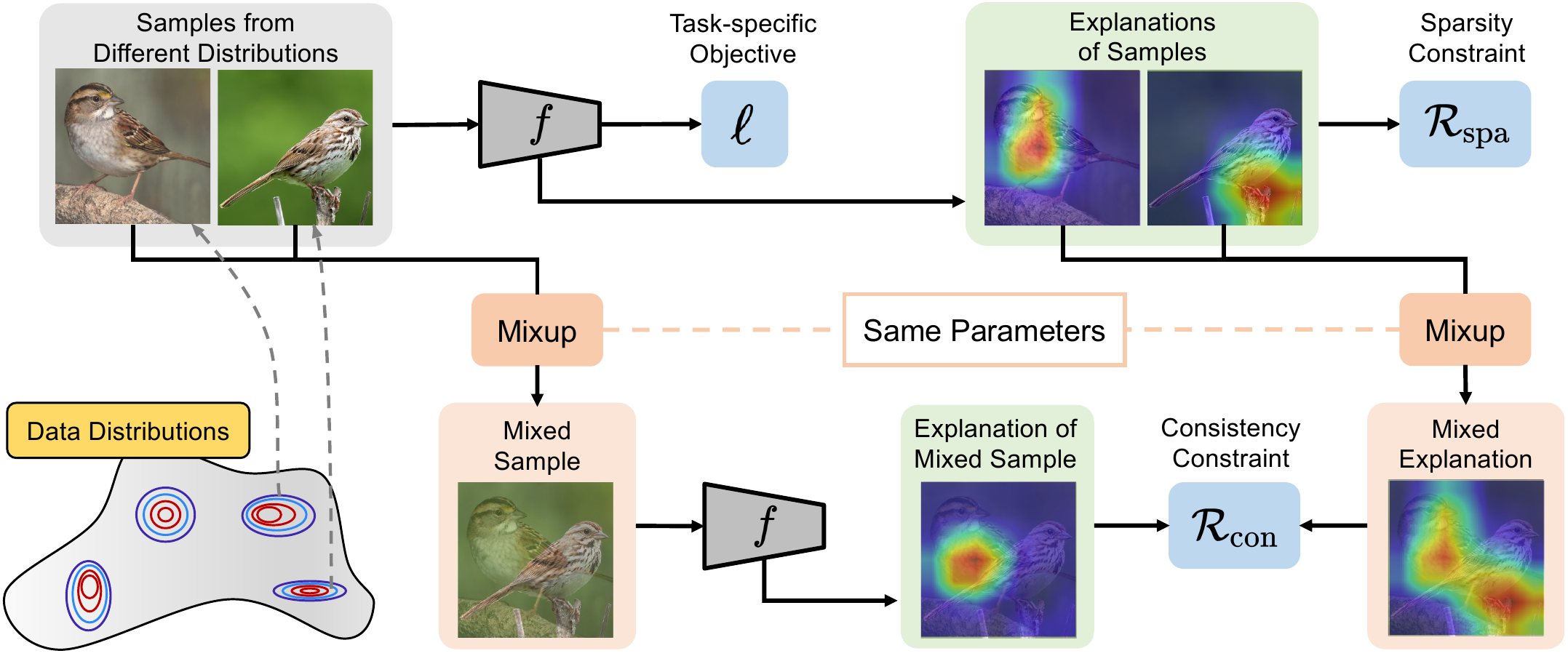}
\caption{
{\bf Overview of the proposed Distributionally Robust Explanations (DRE) method.}
Our method consists of distributional explanation consistency constraint and explanation sparsity constraint.
Firstly, we load a batch of random samples from different distributions and feed them into the model to calculate the standard task-specific objective ({\it e.g.}, Cross-entropy loss for classification tasks).
Then, we calculate the explanations of each sample {\it w.r.t.} their predictions and constrain the explanation sparsity.
Next, we pair up the samples with the same prediction but from different distributions, mixing up both samples and explanations using the same parameters.
We feed the mixed samples into the model and calculate the explanations {\it w.r.t.} their shared predictions.
Finally, we constrain the consistency between the explanation of mixed samples and the mixed explanations.
This figure is best viewed in color.
}
\label{fig:overview}
\end{figure*}

\subsection{Distributionally Robust Explanations}
Question: {\it How to develop robust explanations against out-of-distribution data?}

We started by providing the formulation of the problem, then introduce a new framework {\it Distributionally Robust Explanations} (DRE) for explanation learning.

\subsubsection{Problem formulation}
\label{sec:problem_formulation}
The objective of typical supervised learning is to learn a predictor $f \in \mathcal{F}$ such that $f(x)\rightarrow y$ for any $(x, y) \sim P(X,Y)$, where $P(X,Y)$ is an unknown joint probability distribution, and $\mathcal{F}$ is a function class that is model-agnostic for a prediction task.
However, in the scenario of {\it out-of-distribution} generalization, one can not sample directly from $P(X,Y)$ due to distributional shifts.
Instead, it is assumed that we can only measure $(X, Y)$ under different environmental conditions $e$ so that data is drawn from a set of groups $\mathcal{E}_\mathrm{all}$ such that $(x, y) \sim P_e(X,Y)$.
For example, in the cancer detection task, the environmental conditions denote the latent factors ({\it e.g.}, data acquisition protocols or equipment manufacturers) that underlie different hospitals.

Informally, this assumption specifies that there should exist a function $G$ that relates the random variables $X$ and $X^e$ via $G(X,e) = X^e$.
Let $\mathcal{E}_\mathrm{train} \subsetneq \mathcal{E}_\mathrm{all}$ be a finite subset of training groups (distributions), given the task-specific objective function $\ell$ and explanation method $g(\cdot)$, our {\it Distributionally Robust Explanations} is equivalent to the following constrained problem:
\begin{equation}
\tag{DRE}
\begin{aligned}
& \mathop{\mathrm{min}}\limits_{f\in \mathcal{F}} \quad \mathcal{R}(f) : = \mathbb{E}_{(x,y)\sim P_\mathrm{train}} [\ell(f(x), y)] \\
& \; \mathrm{s.t.} \quad \;\; g(x) = g(G(x, e)) \quad \forall e \in \mathcal{E}_\mathrm{all}.  
\end{aligned}
\label{eq:DRE}
\end{equation}
Intuitively, we encourage the model to have invariant explanations for a sample under different environmental conditions (distributions) after optimization.

The problem in \ref{eq:DRE} is challenging to solve, since we do not have access to the set of all distributions $\mathcal{E}_\mathrm{all}$ or the underlying distribution $P(X,Y)$, and to the distribution transformation function $G$.
The alternative solutions would be:
(i) acquire the ground truth explanations for all samples;
(ii) obtain the one-to-one mapping between samples from different distributions, such as original sample and its corresponding corrupted ones~\cite{hendrycks2018benchmarking}.
However, as our discussion in Sec.~\ref{sec:relate}, the ground truth explanations and one-to-one mappings are practically unavailable in real-world tasks.

\subsubsection{Distributional Explanation Consistency}
\label{sec:DEM}
We address these challenges by explicitly designing the {\it distributional explanation consistency}.
The key idea is, inspired by self-supervised learning, leveraging the mixed explanation to provide supervisory signals for the learning of explanations.
Specifically, we first leverage the {\it distributional mixup} to achieve a simple but effective inter-distributional transformation. 
The {\it mixup}~\cite{zhang2018mixup} methods have been empirically shown to substantially improve performance and robustness to adversarial noise~\cite{qiao2021uncertainty, xu2020adversarial}.
In contrast to original {\it mixup} that mixes random pairs of samples and labels, we mix up samples with the same ground truth but from different distributions.
Denote $(\mathrm{\bf x}_e, \mathrm{\bf x}_{e'})$ as random pairs of training samples with the same ground truth $\mathrm{\bf y}$ but from different distributions, then our {\it mixup} operator can be defined as:
\begin{equation}
  \mathcal{M} (\mathrm{\bf x}_e, \mathrm{\bf x}_{e'}) = \tau  \mathrm{\bf x}_e + (1-\tau  )\mathrm{\bf x}_{e'}
  \label{eq:input_mixup}
\end{equation}
where $\tau \sim \mathrm {Beta}(\alpha, \alpha)$ and the {\it mixup} hyper-parameter $\alpha \in (0, +\infty)$ controls the interpolation strength, in practice we use $\alpha=0.2$.
Secondly, we {\it mixup} the explanations of the samples using the same parameters, namely:
\begin{equation}
  \mathcal{M} (g(\mathrm{\bf x}_e), g(\mathrm{\bf x}_{e'})) = \tau  g(\mathrm{\bf x}_e) + (1-\tau  )g(\mathrm{\bf x}_{e'})
  \label{eq:exp_mixup}
\end{equation}
Thirdly, we feed the mixed samples into the model to calculate the explanations $g(\mathcal{M}(\mathrm{\bf x}_e, \mathrm{\bf x}_{e'}))$.
Lastly, denote $\mathcal{D}$ as an arbitrary discrepancy metric, for example, $\ell_1$ distance and KL-divergence~\cite{kullback1951information}, for all $e,e' \in \mathcal{E}_\mathrm{train}$, we leverage the consistency between the mixed explanation and the explanation of the mixed sample as an alternative of~\ref{eq:DRE}:
\begin{equation}
\mathop{\mathrm{min}}\limits_{f\in \mathcal{F}} \mathcal{R}(f) \;
\mathrm{s.t.} \; \mathcal{D} [g(\mathcal{M}(\mathrm{\bf x}_e, \mathrm{\bf x}_{e'})), \mathcal{M}(g(\mathrm{\bf x}_e), g(\mathrm{\bf x}_{e'}))] \le \epsilon 
\label{eq:L_con}
\end{equation}
Intuitively, the mixed explanation serves as the pseudo label to guide the learning of the explanation for the mixed sample.
Note that $g(\cdot)$ is not restrictive, including any gradient-based explanation methods, such as Grad-CAM~\cite{selvaraju2017grad} and Input Gradient~\cite{simonyan2013deep}.
We leverage Karush–Kuhn–Tucker conditions~\cite{boyd2004convex} and introduce a Lagrange multiplier $\lambda$ to convert the constrained problem in Eq.~\ref{eq:L_con} into its unconstrained counterpart:
\begin{equation}
\begin{aligned}
\mathop{\mathrm{min}}\limits_{f\in \mathcal{F}} \{& \mathcal{R}_\mathrm{con} (f) := \; \mathbb{E}_{(x,y)\sim P_\mathrm{train}} [\ell(f(x), y)] \\
                                                  & + \lambda \mathcal{D} [g(\mathcal{M}(\mathrm{\bf x}_e, \mathrm{\bf x}_{e'})), \mathcal{M}(g(\mathrm{\bf x}_e), g(\mathrm{\bf x}_{e'}))]\}
\end{aligned}
\label{eq:L_con_lagrange}
\end{equation}

{\bf Explanation Regularization}
Our empirical results (Tab.~\ref{tab:ablation}) show that recklessly optimizing Eq.~\ref{eq:L_con_lagrange} could easily fall into a local minimum.
For example, explanations that evenly attribute the prediction to all features in the sample would be a trivial solution to satisfy the constraint.
To address this problem, we propose to further regularize the $\ell_1$-norm of the explanations. 
Let $\gamma$ be a dual variable, our overall objective is formulated as follows:
\begin{equation}
\mathop{\mathrm{min}}\limits_{f\in \mathcal{F}} \{\mathcal{R}(f) := \mathcal{R}_\mathrm{con} (f)
+ \gamma [|| g(\mathrm{\bf x}_e) ||_{1} + || g(\mathrm{\bf x}_{e'}) ||_{1}] \}
\label{eq:L_reg}
\end{equation}

The entire training pipeline is summarized in Alg.~\ref{alg:DEC}.
Our {\it Distributional Robust Explanations} has the following merits.
First, in contrast to existing works that rely on expensive explanation annotations~\cite{rieger2020interpretations, stammer2021right} or one-to-one mapping between image transforms~\cite{guo2019visual, wang2020self, pillai2022consistent}, the proposed method provides a supervisory signal for the explanation learning in general distributional shifts.
Second, the proposed method fully utilizes the inter-distribution information which guarantees distributional invariance in a more continuous latent space.
Third, the proposed method does not involve additional parameters, it can be considered as an add-on module for training a robust model without changing the model architecture.
This enables DRE for robust explanation and prediction against distributional shifts as shown in Sec.~\ref{sec:exp}.

\begin{algorithm}[t]
\caption{The proposed Distributionally Robust Explanations (DRE).}
\LinesNumbered
\label{alg:DEC}
\KwIn{Data of $\mathcal{E}_\mathrm{train}$; Step size $\eta$}
\KwOut{Learned model parameters $\theta$}
\While{not converged}{
     Sample $(\mathrm{\bf x}_e, \mathrm{\bf y})\sim P_e(X,Y)$  $\; \forall e \in \mathcal{E}_\mathrm{train}$ \\
     Sample $(\mathrm{\bf x}_{e'}, \mathrm{\bf y})\sim P_{e'}(X,Y)$  $\; \forall e' \in \mathcal{E}_\mathrm{train}$ \\
     Calculate $\mathcal{R}(f)$ via Eq.~\ref{eq:L_reg} \\
     Update $\theta$ via $\theta^{t+1} = \theta^{t} - \eta^{t}\nabla \mathcal{R}(f) $
}
\end{algorithm}

\subsection{Improve Model's Generalization Capability}
Question: {\it Can robust explanations benefit the model's generalization capability?}

We empirically evaluate this problem on two out-of-distribution generalization benchmark image datasets ({\it Terra Incognita}~\cite{beery2018recognition} and {\it VLCS}~\cite{fang2013unbiased}) and a scientific tabular dataset ({\it Urban Land}~\cite{gao2020mapping}).
Tab.~\ref{tab:question_3} shows the results of prediction performance on OOD data.
For image datasets, on average of each distribution as the testing set, our method outperforms the ERM~\cite{vapnik1999nature} results by 6.9\% and 2.0\% in terms of prediction accuracy.
For scientific tabular data, on average of each continental region as the testing set, our method significantly outperforms the baseline results by 18.5\% in terms of the prediction residual.

Therefore, {\it the robust explanations against OOD data can benefit the model's generalization capability}.
Recklessly minimizing training error would leads machines to absorb all the correlations found in training data~\cite{arjovsky2019invariant}.
We argue that our robust explanations alleviate the excessive reliance of the model on {\it spurious correlations}.
Our method constrains the model to rely on invariant causal correlations and leads to better generalization capability.

\begin{table}[t]
\centering
\begin{tabular}{llc}
\toprule[1pt]
Dataset                          & Method     & Evaluation* \\ \toprule[1pt]
\multirow{2}{*}{Terra Incognita~\cite{beery2018recognition}} & ERM        & 46.1\%      \\
                                 & DRE (ours) & {\bf 53.0\%}     \\ \toprule[0.5pt]
\multirow{2}{*}{VLCS~\cite{fang2013unbiased}}            & ERM        & 77.5\%      \\
                                 & DRE (ours) & {\bf 79.5\%}      \\ \toprule[0.5pt]
\multirow{2}{*}{Urban Land~\cite{gao2020mapping}}      & ERM        & 9.70e-4     \\
                                 & DRE (ours) & {\bf 7.91e-4}     \\ \bottomrule[1pt]
\end{tabular}
\caption{
Evaluation of predictive performance on {\it out-of-distribution} data.
The results are on the average of each distribution as the testing set.
Note that our method outperforms ERM on {\it Terra Incogenita}, {\it VLCS}, and {\it Urban Land}.
*The reported evaluation scores are Accuracy ({\it Terra Incogenita}, {\it VLCS}) and Prediction Residual ({\it Urban Land}).
}
\label{tab:question_3}
\end{table}

%% file: Section_Camera_ready/4_Experiments.tex
\begin{table*}[t]
\centering
\resizebox{2.1\columnwidth}{!}{%
\begin{tabular}{llccccccccccc}
\toprule[1.5pt]
                          &            & \multicolumn{5}{c}{Terra Incognita~\cite{beery2018recognition}}            &  & \multicolumn{5}{c}{VLCS~\cite{fang2013unbiased}}                       \\ \cline{3-7} \cline{9-13} 
Metric                    & Method                                       & Loc.100          & Loc.38           & Loc.43           & Loc.46           & Avg.             &  & Caltech101       & LabelMe          & SUN09           
                          & VOC2007          & Avg.              \\ \toprule[1pt]
\multirow{6}{*}{DEC loss $\downarrow$} & ERM~\cite{vapnik1999nature}                  & 1.014            & 0.971            & 0.998            & 1.016            & 1.000            &  & 0.991            & 0.902            & 0.987           
                          & 1.120            & 1.000             \\
                          & GroupDRO~\cite{sagawa2019distributionally}   & 0.929            & 0.996            & 1.072            & 0.969            & 0.992            &  & 0.947            & 0.907            & 1.026           & 0.942            & 0.956             \\
                          & IRM~\cite{arjovsky2019invariant}             & 0.982            & 0.965            & 0.982            & 0.939            & 0.967            &  & \underline{0.918}& \underline{0.874}&\underline{0.947} & 0.928            & \underline{0.918} \\
                          & Mixup~\cite{xu2020adversarial}               & 0.998            & 0.926            & 1.032            & 0.970            & 0.982            &  & 0.942            & 1.014            & 1.072           & 0.947            & 0.995             \\ \cline{2-13} 
                          & CGC~\cite{pillai2022consistent}              & \underline{0.382}& \underline{0.412}& \underline{0.396}& \underline{0.356}& \underline{0.387}&  & 1.000            & 0.915            & 0.992           & \underline{0.918}& 0.956             \\
                          & DRE (ours)                                   & {\bf 0.195}      & {\bf 0.250}      & {\bf 0.213}      & {\bf 0.226}      & {\bf 0.221}      &  & {\bf 0.061}      & {\bf 0.431}      & {\bf 0.421}      & {\bf 0.229}      & {\bf 0.286}       \\ \toprule[1pt]
\multirow{6}{*}{iAUC $\uparrow $}     & ERM~\cite{vapnik1999nature}                  & 0.517            & 0.644            & \underline{0.614}& \underline{0.560}& 0.584            &  & \underline{0.964}& \underline{0.787}& {\bf 0.765}           
                               & \underline{0.758}& \underline{0.819}             \\
                          & GroupDRO~\cite{sagawa2019distributionally}   & {\bf 0.678}      & 0.578            & 0.608            & 0.525            & \underline{0.597}&  & 0.928            & 0.786            & 0.643           & 0.706            & 0.766            \\
                          & IRM~\cite{arjovsky2019invariant}             & 0.489            & \underline{0.651}& 0.438            & 0.500            & 0.520            &  & 0.939            & 0.772            & 0.613           & 0.715            & 0.760             \\
                          & Mixup~\cite{xu2020adversarial}               & 0.535            & 0.471            & 0.488            & 0.450            & 0.486            &  & 0.924            & 0.767            & 0.607           & 0.641            & 0.735             \\ \cline{2-13} 
                          & CGC~\cite{pillai2022consistent}              & 0.238            & 0.250            & 0.380            & 0.322            & 0.298            &  & 0.804            & 0.581            & 0.524           & 0.618            & 0.632             \\
                          & DRE (ours)                                   & \underline{0.651}& {\bf 0.685}      & {\bf 0.633}      & {\bf 0.605}      & {\bf 0.644}      &  & {\bf 0.973}      & {\bf 0.837}      & \underline{0.752}       & {\bf 0.763}          & {\bf 0.840}             \\ \toprule[1pt]
\multirow{6}{*}{Acc (\%) $\uparrow $} & ERM~\cite{vapnik1999nature}                  & 49.8             & 42.1             & \underline{56.9} & 35.7             & 46.1             &  & 97.7             & 64.3             & 73.4            
                          & 74.6             & 77.5              \\
                          & GroupDRO~\cite{sagawa2019distributionally}   & 41.2             & 38.6             & 56.7             & 36.4             & 43.2             &  & 97.3             & 63.4             & 69.5            & 76.7             & 76.7              \\
                          & IRM~\cite{arjovsky2019invariant}             & 54.6             & 39.8             & 56.2             & 39.6             & 47.6             &  & {\bf 98.6}       & \underline{64.9} & 73.4            & \underline{77.3} & \underline{78.5}  \\
                          & Mixup~\cite{xu2020adversarial}               & \underline{59.6} & 42.2             & 55.9             & 33.9             & \underline{47.9} &  & \underline{98.3} & 64.8             & 72.1            & 74.3             & 77.4              \\ \cline{2-13} 
                          & CGC~\cite{pillai2022consistent}              & 51.8             & \underline{44.6} & 54.9             & \underline{39.8} & 47.8             &  & 97.1             & 63.2             & \underline{73.6} & 70.6             & 76.1              \\
                          & DRE (ours)                                   & {\bf 64.1}       & {\bf 48.1}       & {\bf 57.1}       & {\bf 42.8}       & {\bf 53.0}       &  & \underline{98.3} & {\bf 65.5}       & {\bf 73.8}      & {\bf 80.2}       & {\bf 79.5}        \\ \bottomrule[1.5pt]
\end{tabular}
}
\caption{
Comparison of the {\it out-of-distribution} explanation and prediction performance on {\it Terra Incognita}~\cite{beery2018recognition} and {\it VLCS}~\cite{fang2013unbiased} datasets.
The models are tested on the specified distribution and trained on all other distributions.
Our results are on the average of three trials of experiments.
We highlight the {\bf best results} and the \underline{second best} results.
Note that the Acc (\%) numbers for ERM~\cite{vapnik1999nature}, GroupDRO~\cite{sagawa2019distributionally}, IRM~\cite{arjovsky2019invariant}, and Mixup~\cite{xu2020adversarial} are from Gulrajani and Lopez-Paz~\cite{gulrajani2020search}.
}
\label{tab:terra-vlcs}
\end{table*}

\begin{table*}[t]
\centering
\resizebox{2.1\columnwidth}{!}{%
\begin{tabular}{llcccccccccc}
\toprule[1.5pt]
Metric                                                                                      & Method                             & Africa       & E. Asia       & Europe       & N. Africa       & N. America   & Oceania 
         & Russia       & S. America    & S. Asia        & Avg. \\ \toprule[1pt]
\multirow{2}{*}{DEC loss $\downarrow$}                                                      & ERM~\cite{vapnik1999nature}        & 3.69e-1      & 2.25e-1       & 9.98e-1      & 1.43e+0         & 2.14e+1      & 2.27e+0  
         & 5.56e-1      & 5.58e-1       & 4.54e-1        & 1.00e+0     \\
                                                                                            & DRE (ours)                         & {\bf 2.52e-1}& {\bf 2.11e-1} & {\bf 9.20e-2}& {\bf 1.12e-3}   & {\bf 4.86e-6}& {\bf 7.25e-1}   
         & {\bf 5.17e-6}& {\bf 6.38e-6} & {\bf 1.10e-1}  & {\bf 1.55e-1} \\ \toprule[1pt]
\multirow{2}{*}{SC $\uparrow$}                                                              & ERM~\cite{vapnik1999nature}        & 0.810        &  {\bf 0.779}  & 0.100        & -0.707          & -0.653       & 0.855  
         & -0.856       & -0.837        & -0.504         & -0.113      \\
                                                                                            & DRE (ours)                         & {\bf 0.967}  &  {\bf 0.779}  & {\bf 0.760}  & {\bf 0.135}     & {\bf 0.189}  & {\bf 0.894}   
         & {\bf 0.130}  & {\bf 0.058}   & {\bf 0.324}    & {\bf 0.471}     \\ \toprule[1pt]
\multirow{2}{*}{\begin{tabular}[c]{@{}l@{}}Prediction\\ Residual $\downarrow$\end{tabular}} & ERM~\cite{vapnik1999nature}        & 7.45e-4      & 1.46e-3       & 2.43e-3      & 7.10e-4         & 5.40e-4      & {\bf 1.53e-4} & 8.04e-5      & 3.55e-4       & 2.26e-3        & 9.70e-4     \\
                                                                                            & DRE (ours)                         & {\bf 6.58e-4}& {\bf 1.16e-3} & {\bf 1.95e-3}& {\bf 5.74e-4}   & {\bf 4.14e-4}& 1.56e-4 
         & {\bf 6.34e-5}& {\bf 2.94e-4} & {\bf 1.85e-3}  & {\bf 7.91e-4}     \\ \bottomrule[1.5pt]
\end{tabular}
}
\caption{
Comparison of the {\it out-of-distribution} explanation and prediction performance for short-term urbanization estimation (2000–2010) on the {\it Urban Land}~\cite{gao2020mapping} dataset.
Our results are on the average of three trials of experiments.
Note that a residual of 0.01 indicates a one-percentage point difference between the estimated and observed built-up land fractions.
}
\label{tab:urban}
\end{table*}

To best validate the performance, we conduct a series of experiments to compare our DRE method with existing methods.
The experimental results prove that our method achieves superior explanation and prediction robustness against {\it out-of-distribution} data on a wide scope of tasks, including classification and regression tasks on image and scientific tabular data.

\subsection{Datasets and Implementation Details}
{\bf Datasets.}
We validate our method on two OOD generalization benchmark image datasets for classification and a scientific tabular dataset for regression.
(1) {\it Terra Incognita}~\cite{beery2018recognition} ($\approx$ 11K images, 10 classes) consists of four sub-datasets: Location 100, Location 38, Location 43, and Location 46.
Each sub-dataset indicates a camera trap location in the wild and can be viewed as a different distribution.
Each image in these datasets contains one single animal category ({\it e.g.,} coyote) with different illumination, backgrounds, perspective, etc.
(2) {\it VLCS}~\cite{fang2013unbiased} ($\approx$ 25K images, 5 classes) consists of four sub-datasets: Caltech101~\cite{FeiFei2004LearningGV}, LabelMe~\cite{russell2008labelme}, SUN09~\cite{choi2010exploiting}, and VOC2007~\cite{pascal-voc-2007}.
Each sub-dataset can be viewed as a different distribution.
Each image in these datasets contains one single image category ({\it e.g.,} car) with different styles and backgrounds.
(3) {\it Global National Total Amounts of Urban Land, SSP-Consistent Projections and Base Year, v1 (2000 - 2100)}~\cite{gao2020mapping} (hereinafter referred to as {\it Urban Land}).
The dataset is used for urban land prediction, and the global land area has been divided into 997,022 grid cells.
Each grid cell contains nine topographic, population, and historical urban fraction attributes.
The task is to predict the urban fraction in the year 2010.
The world is been divided into nine continental regions, each region can be viewed as a different distribution.

{\bf Implementation details.}
For all datasets, we alternately leave one distribution out as the testing set.
We split the data from each training distribution into 80\% and 20\% splits, and use the larger splits for training, the smaller splits for validation and model selection.
All models are trained using Adam~\cite{kingma2014adam} optimizer for 5,000 steps.
(1) For both of the image datasets, following the setup in~\cite{gulrajani2020search}, we use a ResNet50~\cite{he2016deep} model pretrained on ImageNet~\cite{deng2009imagenet} and fine-tune.
We freeze all batch normalization layers before fine-tuning and insert a dropout layer before the final linear layer.
We crop the images of random size and aspect ratio, resizing to 224 $\times $ 224 pixels, random horizontal flips, random color jitter, grayscaling the image with 10\% probability, and normalization using the ImageNet channel statistics.
We use learning rate=5e-5 and batch size=16.
(2) For scientific data, following the setup in~\cite{li2021deep}, we transform the tabular data into image-like data with each grid cell as a pixel.
With each land pixel as the center pixel, we densely sampled 997,022 images with size 16 $\times$ 16.
We use a standard U-Net~\cite{ronneberger2015u} model training from scratch.
The images have been augmented by random vertical and horizontal flips, rotation by $90^{\circ}$, $180^{\circ}$, $270^{\circ}$.
We use learning rate=1e-4 and batch size=256.

\begin{figure*}[t]
\centering
\includegraphics[width=1.0\linewidth]{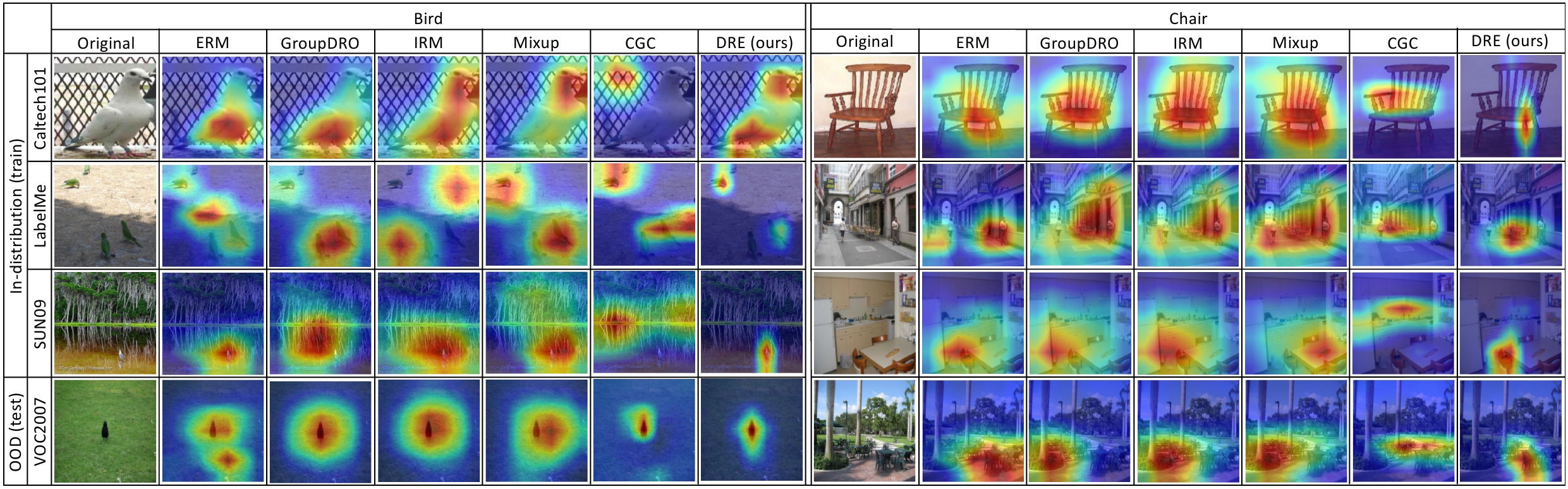}
\caption{
Grad-CAM explanations for images from {\it Bird} (left) and {\it Chair} (right) classes in {\it VLCS} dataset.
The model trained via existing methods, such as ERM~\cite{vapnik1999nature}, Mixup~\cite{xu2020adversarial}, and CGC~\cite{pillai2022consistent}, not only focuses on the objects, but also distribution-specific associations, it getting even severe on OOD data.
On the contrary, our model alleviates the reliance on {\it spurious correlations} ({\it e.g.}, background pixels), and makes consistent explanations on OOD data.
This figure is best viewed in color.
}
\label{fig:exp-vlcs}
\end{figure*}

\subsection{Evaluation Metrics}
\label{sec:metric}
{\bf Distributional explanation consistency (DEC).}
We evaluate the explanations using the same explanation consistency we used for our model training.
Note that although we optimized the in-distribution explanation consistency, the explanation consistency on {\it out-of-distribution} data is still an important metric to evaluate our goal.
We mix up the OOD samples and their explanations with in-distribution samples and their explanations to calculate the DEC loss.
For better comparison, we set the average DEC loss for the model trained via standard ERM~\cite{vapnik1999nature} as 1.0 and rescale others correspondingly.
We expect the DEC loss to be lower for the robust explanations against {\it out-of-distribution} data.

{\bf Explanation fidelity (iAUC).}
This metric measures an increase in the probability of the predictive score as more and more pixels are introduced in a descending order of importance, where the importance is obtained from the explanation~\cite{Petsiuk2018rise, 10.5555/3491440.3491857}.
We measure the area under the insertion curve (iAUC), which accumulated the probability increase while gradually inserting the features from the original input into a reference input.
The reference input we used is an initially blurred canvas, since the blurring takes away most of the finer details of an image without exposing it to sharp edges which might introduce spurious correlations.
To compare between models, we normalize the logit of the correct class to 1.
We expect the iAUC score to be higher for a robust explanation against OOD data.

{\bf Scientific consistency (SC).}
We introduce this metric for the explanations of scientific data.
Scientific consistency means the explanations obtained are plausible and consistent with existing scientific principles~\cite{roscher2020explainable}.
We leverage explanations of domain expert verified models as the ground truth domain knowledge. 
We use it as a posteriori by cosine similarity between feature importance obtained via explanations and domain knowledge.
We expect the SC score to be higher for a robust explanation against distributional shifts.

\subsection{Evaluation on Terra Incognita}
\label{sec:evaluate_terra}
We compare our model with models trained via four representative OOD generalization methods ({\it i.e.}, ERM~\cite{vapnik1999nature}, GroupDRO~\cite{sagawa2019distributionally}, IRM~\cite{arjovsky2019invariant}, and Mixup~\cite{xu2020adversarial}), and a state-of-the-art explanation-guided learning method ({\it i.e.}, CGC~\cite{pillai2022consistent}).
We use Grad-CAM~\cite{selvaraju2017grad} to generate explanations after model training. 
Quantitatively, in Tab.~\ref{tab:terra-vlcs} we report the results of explanation quality and predictive performance on OOD data.
Our model outperforms the four OOD generalization methods and CGC~\cite{pillai2022consistent} with significant improvement on all three metrics.
On average of each distribution as the testing set, our method outperforms the second-best results by 42.9\%, 7.9\%, and 5.1\% in terms of DEC loss, explanation fidelity, and prediction accuracy.
Qualitatively, in Fig.~\ref{fig:intro} we visualize the Grad-CAM explanations of ERM, GroupDRO, IRM, and our model using Location 46 as the testing distribution.
Our explanations corrected the wrongly focused explanations generated by the models of existing methods.
Rather than backgrounds, our explanations are more concentrated on the most discriminative object.
The results demonstrate the superiority of our model on explanation and prediction robustness against OOD data.

\subsection{Evaluation on VLCS}
Following the same settings in Sec.~\ref{sec:evaluate_terra}, we compare our model with ERM~\cite{vapnik1999nature}, GroupDRO~\cite{sagawa2019distributionally}, IRM~\cite{arjovsky2019invariant}, Mixup~\cite{xu2020adversarial}), and CGC~\cite{pillai2022consistent} models.
Quantitatively, as shown in Tab.~\ref{tab:terra-vlcs}, our model outperforms existing methods with significant improvement on all three metrics.
On average of each distribution as the testing set, our method outperforms the second-best results by 68.8\%, 2.6\%, and 1.0\% in terms of DEC loss, explanation fidelity, and prediction accuracy.
Qualitatively, in Fig.~\ref{fig:exp-vlcs} we visualize the Grad-CAM explanation of five existing methods and our method.
Our explanations are more concentrated on the most discriminative features of the object, and significantly alleviate the focus of background pixels on OOD data.
Furthermore, we evaluate the efficiency of our method shown in Tab.~\ref{tab:efficiency}.

\subsection{Evaluation on Urban Land}

We compare our model with models trained via ERM~\cite{vapnik1999nature}.
We leverage the Input Gradient~\cite{simonyan2013deep} to generate explanations after model training, because of its fine-grained resolution and advanced explanation performance on truly continuous inputs~\cite{ross2017right}.
As shown in Tab.~\ref{tab:urban}, our model outperforms the ERM model with significant improvement on all three metrics.
On average of each continental region (distribution) as the testing set, our method outperforms the ERM method by 84.5\%, 29.2\%, and 18.5\% in terms of DEC loss, scientific consistency, and prediction residual.
Note that the higher scientific consistency would further achieve promising and valuable scientific outcomes in the downstream tasks~\cite{roscher2020explainable}.

\begin{table}[t]
\centering
\begin{tabular}{lccc}
\toprule[1pt]
Method                          & \# of params. & Training Time & Acc (\%) \\ \toprule[0.5pt]
ERM~\cite{vapnik1999nature}     & 25.6M         & 18.4min       & 74.6     \\
CGC~\cite{pillai2022consistent} & 25.6M         & 28.8min       & 70.6     \\
DRE (ours)                      & 25.6M         & 33.1min       & {\bf 80.2}     \\ \bottomrule[1pt]
\end{tabular}
\caption{
Efficiency comparison on {\it VLCS}~\cite{fang2013unbiased} dataset using VOC2007~\cite{pascal-voc-2007} as testing distribution.
Our model significantly outperforms existing methods on prediction accuracy with a marginal increase in train time than the CGC~\cite{pillai2022consistent} method, no additional parameters were introduced. 
}
\label{tab:efficiency}
\end{table}

\begin{table}[t]
\centering
\resizebox{1.0\columnwidth}{!}{%
\begin{tabular}{lccc}
\toprule[1pt]
                                                                        & DEC loss $\downarrow$          & iAUC $\uparrow$             & Acc (\%) $\uparrow$ \\ \toprule[0.5pt]
ERM~\cite{vapnik1999nature}                                             & 1.000$\pm$0.02                 & \underline{0.758}$\pm$0.01  & 74.6$\pm$1.3     \\
DRE w/o reg.                                 & 0.909$\pm$0.01                 & 0.756$\pm$0.01              & \underline{78.1}$\pm$0.9     \\
DRE w/o consist.                                   & {\bf 0.364}$\pm$0.03           & 0.698$\pm$0.02              & 71.2$\pm$2.1     \\
DRE (full)                                                              & \underline{0.773}$\pm$0.01     & {\bf 0.772}$\pm$0.01        & {\bf 80.2}$\pm$0.4     \\ \bottomrule[1pt]
\end{tabular}
}
\caption{
Ablation study on {\it VLCS}~\cite{fang2013unbiased} dataset using VOC2007~\cite{pascal-voc-2007} as testing distribution.
Although the variant (w/o explanation regularization) increases the prediction accuracy compared to ERM, it makes limited improvement in DEC loss and slightly dropped on iAUC.
The variant (w/o explanation consistency) has the most significant DEC loss decrease, but it also leads to a severe drop in iAUC and prediction accuracy.
}
\label{tab:ablation}
\end{table}

\subsection{Ablation Study}
In this section, we perform ablation studies to investigate key components of our method proposed in Eq.~\ref{eq:L_con_lagrange} and Eq.~\ref{eq:L_reg}.
The empirical results are shown in Tab.~\ref{tab:ablation}

{\bf Ablation on explanation regularization.}
The variant (w/o explanation regularization) increases the predictive accuracy by 3.5\% compared to ERM, however, the it makes limited improvement in DEC loss and slightly dropped on iAUC.
This indicates limited enhancement of explanation robustness against OOD data, the model might falls into a local minimum to satisfy the {\it distributional explanation consistency} constraint.
For example, an explanation that uniformly attributes the prediction to all features.

{\bf Ablation on distributional explanation consistency.}
The variant (w/o explanation consistency) significantly improved the DEC loss by 63.6\% over the ERM model.
However, the explanation fidelity and prediction accuracy are significantly decreased by 6.0\% and 3.4\%.
This indicates that blindly encouraging the explanation sparsity would hurt the explanation and predictive performance.

\subsection{Generalize to Different Explanation Methods}
\begin{figure}[t]
\centering
   \includegraphics[width=1.0\linewidth]{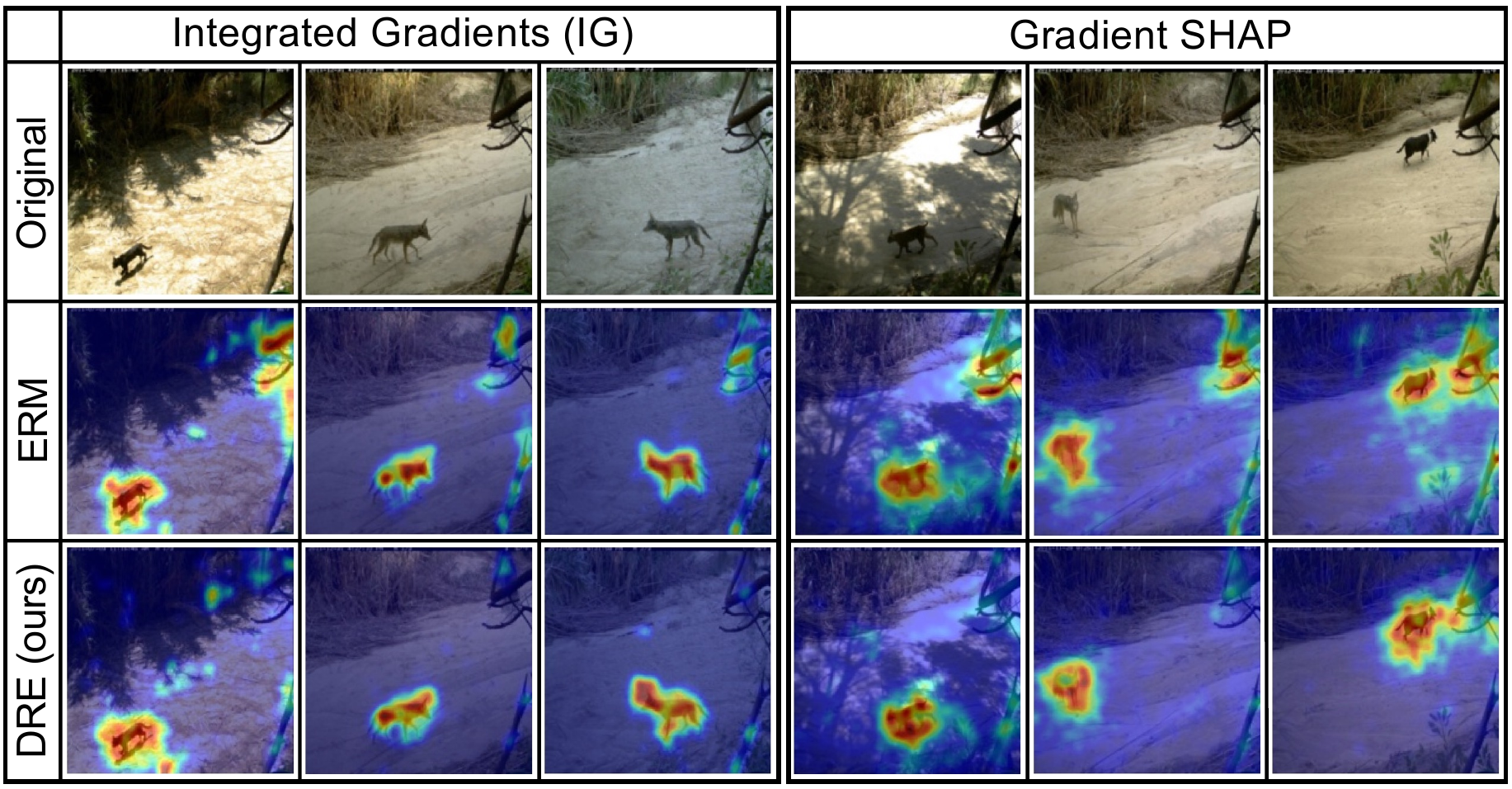}

\caption{
Integrated Gradients (IG)~\cite{sundararajan2017axiomatic} and Gradient SHAP~\cite{lundberg2017unified} saliency maps for OOD data from {\it Terra Incognita} dataset.
Using location 46 as the testing set, for ERM model the saliency maps of both Integrated Gradients (IG)~\cite{sundararajan2017axiomatic} and Gradient SHAP~\cite{lundberg2017unified} methods are excessively focused on background pixels, such as branch and ground.
}
\label{fig:abl-exp}
\end{figure}

In this experiment, we compare the saliency maps of ours and the ERM model.
As shown in Fig.~\ref{fig:abl-exp}, the improved explainability of our model trained by Grad-CAM can be generalized to a variety of data-driven explanation methods.
Using Location 46 in {\it Terra Incognita} as the testing set, for the ERM model, the saliency maps of both Integrated Gradients (IG)~\cite{sundararajan2017axiomatic} and Gradient SHAP~\cite{lundberg2017unified} methods are excessively focused on background pixels, such as branch and ground.
On the contrary, the saliency maps of our model alleviate the reliance on background pixels, and clearly depicts the contour of the object.

%% file: Section_Camera_ready/5_Conclusion.tex
In this paper, we present comprehensive study to show that the data-driven explanations are not robust against out-of-distribution data.
To address this problem, we propose a an end-to-end model-agnostic learning framework Distributionally Robust Explanation (DRE).
The proposed method fully utilizes inter-distribution information to provide supervisory signals for explanation learning without human annotation.
We conduct extensive experiments on wide range of tasks, including the classification and regression tasks on image and scientific tabular data.
Our results demonstrate the superior of our method in terms of explanation and prediction robustness against {\it out-of-distribution} data. 
We anticipate using our robust model with solid justifications in the next efforts for further scientific knowledge discovery.

%% file: Camera_Ready.bbl
\begin{thebibliography}{10}\itemsep=-1pt

\bibitem{adebayo2018sanity}
Julius Adebayo, Justin Gilmer, Michael Muelly, Ian Goodfellow, Moritz Hardt,
  and Been Kim.
\newblock Sanity checks for saliency maps.
\newblock {\em Advances in neural information processing systems}, 31, 2018.

\bibitem{arjovsky2019invariant}
Martin Arjovsky, L{\'e}on Bottou, Ishaan Gulrajani, and David Lopez-Paz.
\newblock Invariant risk minimization.
\newblock {\em arXiv preprint arXiv:1907.02893}, 2019.

\bibitem{bahng2020learning}
Hyojin Bahng, Sanghyuk Chun, Sangdoo Yun, Jaegul Choo, and Seong~Joon Oh.
\newblock Learning de-biased representations with biased representations.
\newblock In {\em International Conference on Machine Learning}, pages
  528--539. PMLR, 2020.

\bibitem{beery2018recognition}
Sara Beery, Grant Van~Horn, and Pietro Perona.
\newblock Recognition in terra incognita.
\newblock In {\em Proceedings of the European conference on computer vision
  (ECCV)}, pages 456--473, 2018.

\bibitem{10.5555/3491440.3491857}
Umang Bhatt, Adrian Weller, and Jos\'{e} M.~F. Moura.
\newblock Evaluating and aggregating feature-based model explanations.
\newblock In {\em Proceedings of IJCAI}, 2021.

\bibitem{boyd2004convex}
Stephen Boyd, Stephen~P Boyd, and Lieven Vandenberghe.
\newblock {\em Convex optimization}.
\newblock Cambridge university press, 2004.

\bibitem{chen2019looks}
Chaofan Chen, Oscar Li, Daniel Tao, Alina Barnett, Cynthia Rudin, and
  Jonathan~K Su.
\newblock This looks like that: deep learning for interpretable image
  recognition.
\newblock {\em Advances in neural information processing systems}, 32, 2019.

\bibitem{chen2019robust}
Jiefeng Chen, Xi Wu, Vaibhav Rastogi, Yingyu Liang, and Somesh Jha.
\newblock Robust attribution regularization.
\newblock {\em Advances in Neural Information Processing Systems}, 32, 2019.

\bibitem{choi2010exploiting}
Myung~Jin Choi, Joseph~J Lim, Antonio Torralba, and Alan~S Willsky.
\newblock Exploiting hierarchical context on a large database of object
  categories.
\newblock In {\em 2010 IEEE computer society conference on computer vision and
  pattern recognition}, pages 129--136. IEEE, 2010.

\bibitem{cugu2022attention}
Ilke Cugu, Massimiliano Mancini, Yanbei Chen, and Zeynep Akata.
\newblock Attention consistency on visual corruptions for single-source domain
  generalization.
\newblock In {\em Proceedings of the IEEE/CVF Conference on Computer Vision and
  Pattern Recognition}, pages 4165--4174, 2022.

\bibitem{deng2009imagenet}
Jia Deng, Wei Dong, Richard Socher, Li-Jia Li, Kai Li, and Li Fei-Fei.
\newblock Imagenet: A large-scale hierarchical image database.
\newblock In {\em 2009 IEEE conference on computer vision and pattern
  recognition}, pages 248--255. Ieee, 2009.

\bibitem{dou2019domain}
Qi Dou, Daniel Coelho~de Castro, Konstantinos Kamnitsas, and Ben Glocker.
\newblock Domain generalization via model-agnostic learning of semantic
  features.
\newblock {\em Advances in Neural Information Processing Systems}, 32, 2019.

\bibitem{pascal-voc-2007}
M. Everingham, L. Van~Gool, C.~K.~I. Williams, J. Winn, and A. Zisserman.
\newblock The {PASCAL} {V}isual {O}bject {C}lasses {C}hallenge 2007 {(VOC2007)}
  {R}esults.
\newblock
  http://www.pascal-network.org/challenges/VOC/voc2007/workshop/index.html.

\bibitem{fang2013unbiased}
Chen Fang, Ye Xu, and Daniel~N Rockmore.
\newblock Unbiased metric learning: On the utilization of multiple datasets and
  web images for softening bias.
\newblock In {\em Proceedings of the IEEE International Conference on Computer
  Vision}, pages 1657--1664, 2013.

\bibitem{FeiFei2004LearningGV}
Li Fei-Fei, Rob Fergus, and Pietro Perona.
\newblock Learning generative visual models from few training examples: An
  incremental bayesian approach tested on 101 object categories.
\newblock {\em Computer Vision and Pattern Recognition Workshop}, 2004.

\bibitem{ganin2016domain}
Yaroslav Ganin, Evgeniya Ustinova, Hana Ajakan, Pascal Germain, Hugo
  Larochelle, Fran{\c{c}}ois Laviolette, Mario Marchand, and Victor Lempitsky.
\newblock Domain-adversarial training of neural networks.
\newblock {\em The journal of machine learning research}, 17(1):2096--2030,
  2016.

\bibitem{gao2020mapping}
Jing Gao and Brian~C O’Neill.
\newblock Mapping global urban land for the 21st century with data-driven
  simulations and shared socioeconomic pathways.
\newblock {\em Nature communications}, 11(1):1--12, 2020.

\bibitem{gulrajani2020search}
Ishaan Gulrajani and David Lopez-Paz.
\newblock In search of lost domain generalization.
\newblock In {\em International Conference on Learning Representations}, 2020.

\bibitem{guo2019visual}
Hao Guo, Kang Zheng, Xiaochuan Fan, Hongkai Yu, and Song Wang.
\newblock Visual attention consistency under image transforms for multi-label
  image classification.
\newblock In {\em Proceedings of the IEEE/CVF conference on computer vision and
  pattern recognition}, pages 729--739, 2019.

\bibitem{han2021explanation}
Tao Han, Wei-Wei Tu, and Yu-Feng Li.
\newblock Explanation consistency training: Facilitating consistency-based
  semi-supervised learning with interpretability.
\newblock In {\em Proceedings of the AAAI conference on artificial
  intelligence}, volume~35, pages 7639--7646, 2021.

\bibitem{he2016deep}
Kaiming He, Xiangyu Zhang, Shaoqing Ren, and Jian Sun.
\newblock Deep residual learning for image recognition.
\newblock In {\em Proceedings of the IEEE conference on computer vision and
  pattern recognition}, pages 770--778, 2016.

\bibitem{hendrycks2018benchmarking}
Dan Hendrycks and Thomas Dietterich.
\newblock Benchmarking neural network robustness to common corruptions and
  perturbations.
\newblock In {\em International Conference on Learning Representations}, 2018.

\bibitem{hind2019ted}
Michael Hind, Dennis Wei, Murray Campbell, Noel~CF Codella, Amit Dhurandhar,
  Aleksandra Mojsilovi{\'c}, Karthikeyan Natesan~Ramamurthy, and Kush~R
  Varshney.
\newblock Ted: Teaching ai to explain its decisions.
\newblock In {\em Proceedings of the 2019 AAAI/ACM Conference on AI, Ethics,
  and Society}, pages 123--129, 2019.

\bibitem{kingma2014adam}
Diederik~P Kingma and Jimmy Ba.
\newblock Adam: A method for stochastic optimization.
\newblock {\em arXiv preprint arXiv:1412.6980}, 2014.

\bibitem{koh2021wilds}
Pang~Wei Koh, Shiori Sagawa, Henrik Marklund, Sang~Michael Xie, Marvin Zhang,
  Akshay Balsubramani, Weihua Hu, Michihiro Yasunaga, Richard~Lanas Phillips,
  Irena Gao, et~al.
\newblock Wilds: A benchmark of in-the-wild distribution shifts.
\newblock In {\em International Conference on Machine Learning}, pages
  5637--5664. PMLR, 2021.

\bibitem{krueger2021out}
David Krueger, Ethan Caballero, Joern-Henrik Jacobsen, Amy Zhang, Jonathan
  Binas, Dinghuai Zhang, Remi Le~Priol, and Aaron Courville.
\newblock Out-of-distribution generalization via risk extrapolation (rex).
\newblock In {\em International Conference on Machine Learning}, pages
  5815--5826. PMLR, 2021.

\bibitem{kullback1951information}
Solomon Kullback and Richard~A Leibler.
\newblock On information and sufficiency.
\newblock {\em The annals of mathematical statistics}, 22(1):79--86, 1951.

\bibitem{lake2017building}
Brenden~M Lake, Tomer~D Ullman, Joshua~B Tenenbaum, and Samuel~J Gershman.
\newblock Building machines that learn and think like people.
\newblock {\em Behavioral and brain sciences}, 40, 2017.

\bibitem{li2018learning}
Da Li, Yongxin Yang, Yi-Zhe Song, and Timothy Hospedales.
\newblock Learning to generalize: Meta-learning for domain generalization.
\newblock In {\em Proceedings of the AAAI conference on artificial
  intelligence}, volume~32, 2018.

\bibitem{li2021deep}
Tang Li, Jing Gao, and Xi Peng.
\newblock Deep learning for spatiotemporal modeling of urbanization.
\newblock {\em Advances in Neural Information Processing Systems Workshops},
  2021.

\bibitem{li2018domain}
Ya Li, Mingming Gong, Xinmei Tian, Tongliang Liu, and Dacheng Tao.
\newblock Domain generalization via conditional invariant representations.
\newblock In {\em Proceedings of the AAAI conference on artificial
  intelligence}, volume~32, 2018.

\bibitem{lundberg2017unified}
Scott~M Lundberg and Su-In Lee.
\newblock A unified approach to interpreting model predictions.
\newblock {\em Advances in neural information processing systems}, 30, 2017.

\bibitem{ma2022multimodal}
Mengmeng Ma, Jian Ren, Long Zhao, Davide Testuggine, and Xi Peng.
\newblock Are multimodal transformers robust to missing modality?
\newblock In {\em Proceedings of the IEEE/CVF Conference on Computer Vision and
  Pattern Recognition}, pages 18177--18186, 2022.

\bibitem{ma2021smil}
Mengmeng Ma, Jian Ren, Long Zhao, Sergey Tulyakov, Cathy Wu, and Xi Peng.
\newblock Smil: Multimodal learning with severely missing modality.
\newblock In {\em Proceedings of the AAAI Conference on Artificial
  Intelligence}, volume~35, pages 2302--2310, 2021.

\bibitem{peng2022out}
Xi Peng, Fengchun Qiao, and Long Zhao.
\newblock Out-of-domain generalization from a single source: An uncertainty
  quantification approach.
\newblock {\em IEEE Transactions on Pattern Analysis and Machine Intelligence},
  2022.

\bibitem{peng2018jointly}
Xi Peng, Zhiqiang Tang, Fei Yang, Rogerio~S Feris, and Dimitris Metaxas.
\newblock Jointly optimize data augmentation and network training: Adversarial
  data augmentation in human pose estimation.
\newblock In {\em Proceedings of the IEEE conference on computer vision and
  pattern recognition}, pages 2226--2234, 2018.

\bibitem{peng2017reconstruction}
Xi Peng, Xiang Yu, Kihyuk Sohn, Dimitris~N Metaxas, and Manmohan Chandraker.
\newblock Reconstruction-based disentanglement for pose-invariant face
  recognition.
\newblock In {\em Proceedings of the IEEE international conference on computer
  vision}, pages 1623--1632, 2017.

\bibitem{Petsiuk2018rise}
Vitali Petsiuk, Abir Das, and Kate Saenko.
\newblock Rise: Randomized input sampling for explanation of black-box models.
\newblock In {\em Proceedings of the British Machine Vision Conference (BMVC)},
  2018.

\bibitem{pillai2022consistent}
Vipin Pillai, Soroush~Abbasi Koohpayegani, Ashley Ouligian, Dennis Fong, and
  Hamed Pirsiavash.
\newblock Consistent explanations by contrastive learning.
\newblock In {\em Proceedings of the IEEE/CVF Conference on Computer Vision and
  Pattern Recognition}, pages 10213--10222, 2022.

\bibitem{qiao2021uncertainty}
Fengchun Qiao and Xi Peng.
\newblock Uncertainty-guided model generalization to unseen domains.
\newblock In {\em Proceedings of the IEEE/CVF Conference on Computer Vision and
  Pattern Recognition}, pages 6790--6800, 2021.

\bibitem{qiao2023topology}
Fengchun Qiao and Xi Peng.
\newblock Topology-aware robust optimization for out-of-distribution
  generalization.
\newblock In {\em Proceedings of the International Conference on Learning
  Representations (ICLR)}, 2023.

\bibitem{qiao2020learning}
Fengchun Qiao, Long Zhao, and Xi Peng.
\newblock Learning to learn single domain generalization.
\newblock In {\em Proceedings of the IEEE/CVF Conference on Computer Vision and
  Pattern Recognition}, pages 12556--12565, 2020.

\bibitem{ribeiro2016should}
Marco~Tulio Ribeiro, Sameer Singh, and Carlos Guestrin.
\newblock " why should i trust you?" explaining the predictions of any
  classifier.
\newblock In {\em Proceedings of the 22nd ACM SIGKDD international conference
  on knowledge discovery and data mining}, pages 1135--1144, 2016.

\bibitem{rieger2020interpretations}
Laura Rieger, Chandan Singh, William Murdoch, and Bin Yu.
\newblock Interpretations are useful: penalizing explanations to align neural
  networks with prior knowledge.
\newblock In {\em International conference on machine learning}, pages
  8116--8126. PMLR, 2020.

\bibitem{ronneberger2015u}
Olaf Ronneberger, Philipp Fischer, and Thomas Brox.
\newblock U-net: Convolutional networks for biomedical image segmentation.
\newblock In {\em International Conference on Medical image computing and
  computer-assisted intervention}, pages 234--241. Springer, 2015.

\bibitem{roscher2020explainable}
Ribana Roscher, Bastian Bohn, Marco~F Duarte, and Jochen Garcke.
\newblock Explainable machine learning for scientific insights and discoveries.
\newblock {\em Ieee Access}, 8:42200--42216, 2020.

\bibitem{ross2017right}
Andrew~Slavin Ross, Michael~C Hughes, and Finale Doshi-Velez.
\newblock Right for the right reasons: training differentiable models by
  constraining their explanations.
\newblock In {\em Proceedings of the 26th International Joint Conference on
  Artificial Intelligence}, pages 2662--2670, 2017.

\bibitem{rudin2019stop}
Cynthia Rudin.
\newblock Stop explaining black box machine learning models for high stakes
  decisions and use interpretable models instead.
\newblock {\em Nature Machine Intelligence}, 1(5):206--215, 2019.

\bibitem{russell2008labelme}
Bryan~C Russell, Antonio Torralba, Kevin~P Murphy, and William~T Freeman.
\newblock Labelme: a database and web-based tool for image annotation.
\newblock {\em International journal of computer vision}, 77(1):157--173, 2008.

\bibitem{sagawa2019distributionally}
Shiori Sagawa, Pang~Wei Koh, Tatsunori~B Hashimoto, and Percy Liang.
\newblock Distributionally robust neural networks.
\newblock In {\em International Conference on Learning Representations}, 2019.

\bibitem{selvaraju2017grad}
Ramprasaath~R Selvaraju, Michael Cogswell, Abhishek Das, Ramakrishna Vedantam,
  Devi Parikh, and Dhruv Batra.
\newblock Grad-cam: Visual explanations from deep networks via gradient-based
  localization.
\newblock In {\em Proceedings of the IEEE international conference on computer
  vision}, pages 618--626, 2017.

\bibitem{shankar2018generalizing}
Shiv Shankar, Vihari Piratla, Soumen Chakrabarti, Siddhartha Chaudhuri, Preethi
  Jyothi, and Sunita Sarawagi.
\newblock Generalizing across domains via cross-gradient training.
\newblock In {\em International Conference on Learning Representations}, 2018.

\bibitem{simonyan2013deep}
Karen Simonyan, Andrea Vedaldi, and Andrew Zisserman.
\newblock Deep inside convolutional networks: Visualising image classification
  models and saliency maps.
\newblock {\em arXiv preprint arXiv:1312.6034}, 2013.

\bibitem{stammer2021right}
Wolfgang Stammer, Patrick Schramowski, and Kristian Kersting.
\newblock Right for the right concept: Revising neuro-symbolic concepts by
  interacting with their explanations.
\newblock In {\em Proceedings of the IEEE/CVF Conference on Computer Vision and
  Pattern Recognition}, pages 3619--3629, 2021.

\bibitem{sundararajan2017axiomatic}
Mukund Sundararajan, Ankur Taly, and Qiqi Yan.
\newblock Axiomatic attribution for deep networks.
\newblock In {\em International conference on machine learning}, pages
  3319--3328. PMLR, 2017.

\bibitem{torralba2011unbiased}
Antonio Torralba and Alexei~A Efros.
\newblock Unbiased look at dataset bias.
\newblock In {\em CVPR 2011}, pages 1521--1528. IEEE, 2011.

\bibitem{vapnik1999nature}
Vladimir Vapnik.
\newblock {\em The nature of statistical learning theory}.
\newblock Springer science \& business media, 1999.

\bibitem{vaswani2017attention}
Ashish Vaswani, Noam Shazeer, Niki Parmar, Jakob Uszkoreit, Llion Jones,
  Aidan~N Gomez, {\L}ukasz Kaiser, and Illia Polosukhin.
\newblock Attention is all you need.
\newblock {\em Advances in neural information processing systems}, 30, 2017.

\bibitem{volpi2018generalizing}
Riccardo Volpi, Hongseok Namkoong, Ozan Sener, John~C Duchi, Vittorio Murino,
  and Silvio Savarese.
\newblock Generalizing to unseen domains via adversarial data augmentation.
\newblock {\em Advances in neural information processing systems}, 31, 2018.

\bibitem{wang2020self}
Yude Wang, Jie Zhang, Meina Kan, Shiguang Shan, and Xilin Chen.
\newblock Self-supervised equivariant attention mechanism for weakly supervised
  semantic segmentation.
\newblock In {\em Proceedings of the IEEE/CVF Conference on Computer Vision and
  Pattern Recognition}, pages 12275--12284, 2020.

\bibitem{wexler2017computer}
Rebecca Wexler.
\newblock When a computer program keeps you in jail.
\newblock {\em The New York Times}, 13, 2017.

\bibitem{xu2020adversarial}
Minghao Xu, Jian Zhang, Bingbing Ni, Teng Li, Chengjie Wang, Qi Tian, and
  Wenjun Zhang.
\newblock Adversarial domain adaptation with domain mixup.
\newblock In {\em Proceedings of the AAAI Conference on Artificial
  Intelligence}, volume~34, pages 6502--6509, 2020.

\bibitem{zhang2018mixup}
Hongyi Zhang, Moustapha Cisse, Yann~N Dauphin, and David Lopez-Paz.
\newblock mixup: Beyond empirical risk minimization.
\newblock In {\em International Conference on Learning Representations}, 2018.

\bibitem{zhou2022feature}
Yilun Zhou, Serena Booth, Marco~Tulio Ribeiro, and Julie Shah.
\newblock Do feature attribution methods correctly attribute features?
\newblock In {\em Proceedings of the AAAI Conference on Artificial
  Intelligence}, volume~36, pages 9623--9633, 2022.

\end{thebibliography}
